\documentclass{article}

\PassOptionsToPackage{numbers, compress}{natbib}
\usepackage[preprint]{neurips_2020}
\usepackage[utf8]{inputenc} % allow utf-8 input
\usepackage[T1]{fontenc}    % use 8-bit T1 fonts
\usepackage{hyperref}       % hyperlinks
\usepackage{url}            % simple URL typesetting
\usepackage{booktabs}       % professional-quality tables
\usepackage{amsmath}
\usepackage{amsfonts}       % blackboard math symbols
\usepackage{nicefrac}
\usepackage{amssymb}
\usepackage{microtype}      % microtypography
\usepackage{bm}
\usepackage{algorithm}
\usepackage{algorithmic}
\usepackage{graphicx,xcolor}

\def\eq#1{(\ref{#1})}

\def\beginmat{ \left( \begin{array} }
\def\endmat{ \end{array} \right) }
\def\diag{{\rm diag}}
\def\log{{\rm log}}
\def\tr{{\rm tr}}
\def\cond{\, | \,}

\newcommand*\diff{\mathop{}\!\mathrm{d}}
\newcommand{\bx}{\mathbf{x}}

\newcommand{\bz}{\mathbf{z}}

\newcommand{\bR}{\mathbf{R}}

\newcommand{\bD}{\mathbf{D}}
\newcommand{\bI}{\mathbf{I}}

\newcommand{\T}{\intercal}

\newcommand{\E}{\mathbb{E}}
\newcommand{\V}{\mathbb{V}}
\newcommand{\N}{\mathcal{N}}

\newcommand{\bvarepsilon}{\boldsymbol\varepsilon}
\newcommand{\bbeta}{\boldsymbol\beta}
\newcommand{\bsigma}{\boldsymbol\sigma}

\newcommand{\btheta}{\boldsymbol\theta}

\newcommand{\bphi}{\boldsymbol\phi}

\newcommand{\bmu}{\boldsymbol\mu}

\newcommand{\bSigma}{\boldsymbol\Sigma}

\newtheorem{theorem}{Theorem}
\title{Generalizing Variational Autoencoders with Hierarchical Empirical Bayes}

\author{%
  Wei Cheng \quad Gregory Darnell \quad Sohini Ramachandran \quad Lorin Crawford\thanks{Corresponding Email: \url{lorin_crawford@brown.edu}}\\
  Brown University\\
  Providence, RI 02912 \\
}

\begin{document}

\maketitle

\begin{abstract}

% mention mutual information?
% aggregated posterior leads to dropping the MI term
% leads to prevent overregularization problem
% Our experiments show we similar to WAE, dropping the MI term leads to prevent collapse
% wei
% we retain VI structure of the ELBO loss, but we share the merits of the WAE in an aggregated posterior
% we posit that the level of noise is directly related to the penalty on the MI term in the ELBO loss
% MI is penalized with small amount of noise. leads to closer to WAE but smoother latent space

% throw WAE under the boat! non-smooth latent space

Variational Autoencoders (VAEs) have experienced recent success as data-generating models by using simple architectures that do not require significant fine-tuning of hyperparameters. However, VAEs are known to suffer from over-regularization which can lead to failure to escape local maxima. This phenomenon, known as posterior collapse, prevents learning a meaningful latent encoding of the data. Recent methods have mitigated this issue by deterministically moment-matching an aggregated posterior distribution to an aggregate prior.
However, abandoning a probabilistic framework (and thus relying on point estimates) can both lead to a discontinuous latent space and generate unrealistic samples. Here we present Hierarchical Empirical Bayes Autoencoder (HEBAE), a computationally stable framework for probabilistic generative models. Our key contributions are two-fold. First, we make gains by placing a hierarchical prior over the encoding distribution, enabling us to adaptively balance the trade-off between minimizing the reconstruction loss function and avoiding over-regularization. Second, we show that assuming a general dependency structure between variables in the latent space produces better convergence onto the mean-field assumption for improved posterior inference. Overall, HEBAE is more robust to a wide-range of hyperparameter initializations than an analogous VAE. Using data from MNIST and CelebA, we illustrate the ability of HEBAE to generate higher quality samples based on FID score than existing autoencoder-based approaches.

\end{abstract}

\section{Introduction} \label{intro}

 Generative modeling has achieved tremendous success in recent years by enabling unsupervised learning of different data distributions, as well as interpretation of data via low-dimensional representations. There two popular approaches in this space: Generative Adversarial Networks (GANs) \cite{goodfellow2014generative} and Variational Autoencoders (VAEs) \cite{kingma2013autoencoding}. In GANs, one plays a min-max game between a discriminator and a generator where the generator is trained to produce high quality samples that fool the discriminator. GANs suffer from a lack of theoretical support that produces problems like ``mode collapse" and makes training difficult \cite{tolstikhin2017wasserstein, ghosh2019variational, dai2019diagnosing}. VAEs, on the other hand, use the neural network architecture of autoencoders and further draw on variational inference. Instead of simply encoding input features into isolated variables in the latent (or hidden) space and reconstructing them using decoders, VAEs further impose a standard normal prior distribution over the latent variables. This prior smooths the regularized latent space during training, thereby enabling the generation of meaningful samples. VAEs leverage well-established theory and are easier to train than GANs. However, VAEs have been know to generate lower quality samples than GANs, and can also result in over-regularization problems such as posterior collapse \cite{bowman2015generating, ghosh2019variational, Phuong2018TheMA, razavi2019preventing}. Previous studies have used manual tuning of hyperparameters to prevent over-regularization and have made various attempts to improve samples quality \cite{tolstikhin2017wasserstein,bowman2015generating, ghosh2019variational, Phuong2018TheMA, razavi2019preventing}. One successful effort referred to as Wassertein Autoencoders (WAEs) \cite{tolstikhin2017wasserstein} offer an alternative framework that remedy the issues in VAEs. WAEs abandon the variational inference framework and minimize the penalized distance between the observed data and target distribution. Though it was shown in the original study that WAEs have the potential to generate better quality samples than VAEs, it has also been reported that WAEs are not robust to hyper-parameter settings \cite{dai2019diagnosing}. WAEs are specified under two versions: one that is based on maximum mean discrepency (WAE-MMD) and the other uses GANs (WAE-GAN). Since the WAE-GAN adopts the potentially unstable adversarial learning, we will treat the WAE-MMD as a baseline throughout the paper.
 
Here we present the Hierarchical Empirical Bayes Autoencoder (HEBAE), a new method that can build generative models and overcomes the challenges of VAEs and WAEs. In designing HEBAE, we connect its theoretical underpinnings with previous efforts like WAEs. We also provide theoretical analyses of the over-regularization problem in VAEs and demonstrate how our method overcomes this problem.  We empirically assess the performance of HEBAE on two real-world image datasets (MNIST and CelebA) and show that HEBAE is both easier to train and capable of generating higher quality samples than competing approaches. All code and data are freely available online at \url{https://github.com/ramachandran-lab/HEBAE}.

\section{Related Work}

\subsection{Variational Autoencoders (VAEs)}

We begin with reviewing the modeling assumptions underlying the variational autoencoder (VAE) framework.
%In this section, we will assume that we have a have a batch of $m$ training examples $\bx_{1:m} = (\bx_1,\ldots,\bx_m)$, each with a set of input features. 
An autoencoder has two key components: an encoder which compresses the original inputs $\bx_i$ to a lower $k$-dimensional latent variable $\bz$, and a decoder which takes those latent variables $\bz$ and attempts to reconstruct the original data (often denoted by $\bx_i^{\prime}$). Intuitively, a successfully trained model aims to minimize the loss function $\sum_i \|\bx_i - \bx_i^{\prime}\|^2$. The VAE framework can be viewed as a probabilistic version of an autoencoder where the posterior distribution of the latent variable $\bz$ is imposed to match a prior distribution $p_{\theta}(\bz)$. By matching the targeted distribution, VAEs can learn a smooth latent space such that it will not just encode isolated data points but produces a generative model over the underlying latent variables \cite{kingma2013autoencoding}. Typical VAEs assume a standard Gaussian prior distribution for $p_{\theta}(\bz) = \N(\bm{0},\bI)$ with zero mean vector and an independent variance-covariance structure between the latent variables. More specifically, the encoder portion of a VAE aims to find a ``best'' approximation to the prior $q_{\phi}(\bz\cond\bx)$ based on the data and a set of free parameters $\phi$. The decoder then aims to construct a likelihood $p_{\theta}(\bx\cond\bz)$ conditioned on the latent variables $\bz$. The goal of variational inference is to maximize the marginal log-likelihood for each example $\bx_i$ in the batch, which takes the following form
\begin{equation}
\log\,p_{\theta}(\bx_i) = \text{KL}(q_{\phi}(\bz\cond\bx_i)\,\|\, p_{\theta}(\bz \cond \bx_i)) + \E_{q_{\phi}(\bz\cond\bx_i)}[\log\,p_{\theta}(\bx_i, \bz)-\log\,q_{\phi}(\bz\cond\bx_i)]\label{logp}
\end{equation}
where the first term measures the Kullback-Leibler (KL) divergence between the approximate and true posterior distribution of the latent $\bz$ given the data $\bx_i$.

Using Jensen's equality, one can  formulate a lower bound to the marginal log-likelihood in Eq.~\eq{logp}, and then iteratively adjust the free parameters $\phi$ so that this bound becomes as tight as possible. It can be shown that finding the ``best'' approximation in the encoder amounts to finding the free parameters $\phi$ that minimizes the KL divergence between $q_{\phi}(\bz\cond\bx)$ and $p_{\theta}(\bz)$ \cite{kingma2013autoencoding}. Taking a stochastic gradient variational Bayes (SGVB) \cite{kingma2013autoencoding} view on VAEs yields the following general expression for the lower bound of the log-likelihood
\begin{equation}
\begin{aligned}
\log\,p_{\theta}(\bx_i) \geq \mathcal{L}_{\text{VAE}}(\theta, \phi; \bx_i)
&\approx \frac{1}{L}\sum_{l=1}^L\log\,p_{\theta}(\bx_i\cond\bz^{(l)})-\,\text{KL}(q_{\phi}(\bz\cond\bx_i)\,\|\, p_{\theta}(\bz))\label{elbo}
\end{aligned}
\end{equation}
 The first term on the right hand side of Eq.~\eq{elbo} is normally referred to as the ``reconstruction loss'' \cite{kingma2013autoencoding, ghosh2019variational, tolstikhin2017wasserstein} and resembles the regular loss function in autoencoders. The second term on the right hand side of Eq.~\eq{elbo} can then be viewed as a ``regularized loss'' function where the variational posterior distribution $q_\phi(\bz\cond\bx_i)$ is being adjusted to approximate the prior $p_\theta(\bz)$ \cite{kingma2013autoencoding, ghosh2019variational, tolstikhin2017wasserstein}. %Note that the KL divergence has a closed form solution when measuring the divergence between two Gaussian distributions. 
Lastly, we use $\bz^{(l)}$ to denote an empirically sampled latent variable using the re-parameterization trick where 
\begin{equation}
\bz^{(l)} = \bmu(\bx_i) + \bsigma(\bx_i) \odot \bvarepsilon^{(l)}, \quad \quad \bvarepsilon\sim\N(\bm{0},\bI)\label{trick}
\end{equation}
To compute an $L$ number of realizations of $\bz$, the encoder outputs a $k$-dimensional mean vector $\bmu(\bx_i) = [\mu(x_{i1}),\ldots,\mu(x_{ik})]$ and a $k$-dimensional vector of standard deviations $\bsigma(\bx_i) = [\sigma(x_{i1}),\ldots,\sigma(x_{ik})]$ for each $\bx_i$. By sampling random noise to determine each $\bz^{(1)},\ldots,\bz^{(L)}$, we are able to analytically compute the reconstruction loss and the regularized loss in Eq.~\eq{elbo}. 

\subsection{Challenges and Disadvantages of the VAE Framework}

There are two key challenges and disadvantages with the VAE framework. First, in practice to train a VAE, one has to balance between minimization of the reconstruction loss and the regularized loss via the KL divergence \cite{bauer2018resampled, dai2019diagnosing}. Concentrating effort on the latter can result in over-regularization of the approximate posterior distribution $q_{\phi}(\bz\cond\bx_i)$ ~\cite{ghosh2019variational}. A  related issue is posterior collapse \cite{razavi2019preventing,oord2017neural,he2019lagging}, where a local optimum of the VAE objective is obtained such that $q_{\phi}(\bz\cond \bx_i) = p_{\theta}(\bz)$ and $\text{KL}(q_{\phi}(\bz\cond\bx_i)\,\|\, p_{\theta}(\bz))\rightarrow 0$. 
%In other words, while part of the goal is to minimize the KL divergence, one has to be careful to avoid actually achieving the optimum where KL is exactly 0. To see this, remember that the regularized loss imposes that the encoder outputs a zero mean vector $\bmu(\bx_i) = \bm{0}$ and a vector of ones for the standard deviation $\bsigma(\bx_i) = \bm{1}$ regardless of the input $\bx_i$, while the reconstruction loss imposes that the decoder reconstructs $\bx_i$ from $\bmu(\bx_i)$ and $\bsigma(\bx_i)$. 
In an extreme case, if the KL divergence dictates that the encoder always outputs $\bmu(\bx_i) = \bm{0}$ and $\bsigma(\bx_i) = \bm{1}$ for all samples, the decoder will face the impossible task of reconstructing different samples from completely random noise $\bz^{(l)} = \bvarepsilon^{(l)}$. To effectively balance this trade-off, VAEs require manually fine-tuning the weight of the KL component and model-specific hyper-parameters~\cite{bowman2015generating, bauer2018resampled}. Moreover, finding the optimal value of the KL divergence remains an open question \cite{hoffman2016elbo}. Since the optimal trade off is unclear, VAEs often generate low quality samples even when the model is fine-tuned.

The second key disadvantage of the VAE framework lies within the assumption that the variational posterior distribution follows an isotropic Gaussian (e.g., Eq.~\eq{trick}). In this case, the VAE cannot completely guarantee that the inference algorithm will converge onto the standard Gaussian prior if the $k$ latent neurons $\bz = (z_{1},\ldots,z_{k})$ are in fact correlated. To see this, notice that the practical lower bound for the joint log-likelihood across all $m$ examples in the training batch is given as 
\begin{equation}
\begin{aligned}
\mathcal{L}_{\text{VAE}}(\btheta, \bphi; \bx_{1:m}) &\approx \frac{1}{L}\sum_{l=1}^L\sum_{i=1}^{m}\log\,p_{\theta}(\bx_i\cond\bz^{(l)})-\lambda\sum_{i=1}^{m}\sum_{j=1}^{k}\left[\log\,\sigma_{ij}^{2}+1-\sigma_{ij}^{2}-\mu_{ij}^{2}\right] \label{elbo2}
\end{aligned}
\end{equation}
where $\lambda\ge 0$ is a practical weight parameter for the KL term to balance the trade-off with reconstruction \cite{ghosh2019variational, higgins2017beta}; $\mu_{ij}$ and $\sigma_{ij}$ denotes the $j$-th term of $\bmu(\bx_i)$ and $\bsigma(\bx_i)$, and the last term is the closed form of the KL divergence between two Gaussian distributions $q_{\phi}(\bz\cond\bx_i) = \N(\bmu(\bx_i),\bD(\bx_i))$ where $\bD(\bx_i) = \diag(\bsigma(\bx_i))$ and $p_{\theta}(\bz) = \N(\bm{0},\bI)$. While the VAE will impose that each $\sigma^2_{ij} \rightarrow 1$ and $\mu_{ij} \rightarrow 0$, in the event that the latent variables are correlated, the model will not approach an optimal fit without also imposing that the covariance $\V[z_{ij},z_{ij^{\prime}}] = 0$ for every $j \ne j^{\prime}$ combination. 

In this work, we will show that placing a flexible prior distribution over the mean function of the encoder enables us to efficiently achieve the optimal trade-off between the reconstruction and regularization loss function in the VAE framework. In our framework assume a general covariance structure for the latent variables without sacrificing model efficiency. With our hierarchical prior specification, our model results in better matching posteriors and higher quality generated samples. 

%Lastly, a sufficiently tuned VAE does not guarantee that the model will avoid overfitting. This is because the variational inference algorithm imposes that the conditional distribution based on the input data $q_{\phi}(\bz\cond\bx_i)$ conforms to match the target prior distribution $p_\theta(\bz)$. However, since $q_{\phi}(\bz\cond\bx_i)$ is not the distribution that we use for predictions, new images generated from this trained model can be of low quality \textcolor{red}{\textbf{[citation]}}. Instead, one should want the average distribution over all training examples $q_{\phi}^{\text{avg}}(\bz)$ to match the targeted prior.

\subsection{Wasserstein Autoencoders and Maximum Mean Discrepancy}

A major goal of the generative model within autoencoders is to derive a smoothed latent space. Recently, there have been many efforts that aim to improve this portion of the VAE framework. Here, we briefly review one popular approach.
%\paragraph{Adversarial Autoencoder (AAE).} The adversarial autoencoder (AAE) makes use of adversarial learning strategy to impose smoothness over the latent space. Under the AAE framework, one abandons variational inference on the per-sample posterior distribution $q_\phi(\bz\cond\bx_i)$ and instead focuses on imposing that the ``aggregated" posterior $q_\phi(\bz)$ matches the standard normal prior where (for large sample sizes $m$) \cite{DBLP:journals/corr/MakhzaniSJG15}
%\begin{equation}
%\begin{aligned}
%q_\phi(\bz) = \int q_\phi(\bz\cond\bx_i) p_d(\bx_i) %\diff \bx_i&\approx \frac{1}{m}\sum_{i=1}^{m}q_\phi(\bz\cond\bx_i)\\
%&\approx \frac{1}{m}\sum_{i=1}^{m}p_\theta(\bz\cond\bx_i) \approx \int p_\theta(\bz\cond\bx_i) p_d(\bx_i) \diff \bx_i = p_\theta(\bz).
%\label{agg_post}
%\end{aligned}
%\end{equation}
%To apply adversarial learning, one couples a discriminator network to the autoencoder, which requires extra weights that need be trained. Although adversarial learning has been shown to have the ability to successfully derive a smooth latent space and produce high quality images, the lack of theoretical foundations in the framework can lead to serious problems such as mode collapse and the model itself is generally harder to train \cite{ghosh2019variational, dai2019diagnosing}.  
The Wasserstein autoencoder (WAE) starts from an optimal transport point-of-view and aims to force the ``aggregated'' posterior distribution $q(\bz)$ to match the standard normal prior \cite{DBLP:journals/corr/MakhzaniSJG15}. One major difference between the WAE and VAE is that the WAE uses a deterministic mapping function between the training inputs $\bx$ and the latent variables $\bz$. As stated in the original paper \cite{tolstikhin2017wasserstein}, one can write the WAE in terms of the VAE framework by regarding $q_{\phi}(\bz \cond \bx_i)$ as a delta mass function $\delta(\bm{\mu}(\bx_i))$ such that $\bz = \bmu(\bx_i)$. The WAE then imposes that the aggregated posterior $q_{\phi}(\bz)=q_{\phi}(\bm{\mu}(\bx_i))$ matches the standard normal by replacing the KL divergence in the regularized loss function with a maximum mean discrepancy (MMD) distance between $q_{\phi}(\bz)$ and the standard normal prior. Similar to VAE, the WAE does not explicitly penalize the covariance between latent variables and, thus, cannot guarantee independence between features at convergence. 

\subsection{Disadvantages of the WAE Framework.} 

In the next section, we argue that completely abandoning a probabilistic framework and relying solely on deterministic point estimates can lead to a non-smooth latent space and result in a generative model that produces unrealistic samples. Indeed, previous studies have shown that the aim of the WAE is equivalent to maximizing a ``looser'' lower bound \cite{hoffman2016elbo,tolstikhin2017wasserstein}. In this study, we propose a new framework that combines the merits of the WAE and VAE. We retain the probabilistic nature of the VAE so we can smooth out the latent space, and we further regularize the aggregated posterior so we can relax the trade-off in the loss function much like the WAE.

%%%%%%%%%%%%%%%%%%%%%%%%%%%%%%%%%%%%%%%%%%%%%%%%%%%%%%%%%%%%

\section{Proposed Method}

\subsection{Hierarchical Empirical Bayes Autoencoders (HEBAE)}

In this section, we present the Hierarchical Empirical Bayes Autoencoder (HEBAE) framework. For convenience, we will follow previous work and treat each sample index $i$ as a random variable \cite{hoffman2016elbo}. To this end, we will denote distributions as $q_\phi(\bz_i\cond\bx_i)$ and $p_\theta(\bz_i)$ with subscripts corresponding to the $i$th sample. The two key components of the HEBAE framework is that \textit{(i)} it assumes a Gaussian process (GP) prior over the encoder function that takes input features to the compressed latent space $\bmu:\bx_i \rightarrow \bz_i$, and \textit{(ii)} it assumes a general covariance structure between the latent variables. For the first component, we assume a hierarchical prior where $\bmu(\cdot)$ is completely specified by its mean function and positive definite covariance (kernel) function, $\bbeta(\cdot)$ and $\bSigma(\cdot,\cdot)$, respectively. In practice, since we have a finite number of examples in a given training batch, we can take a weight-space view on the Gaussian process and equivalently say 
\begin{equation}
\bmu(\bx_i) \sim \N(\bbeta,\bSigma) \label{gp_prior}
\end{equation}
where the encoder function $\bmu$ is assumed to follow from a multivariate normal distribution with mean vector $\bbeta$ and general variance-covariance structure $\bSigma$. For the second key component in the HEBAE framework, we choose non-isotropic Gaussian distributions as our approximating posterior 
\begin{equation}
q_\phi(\bz_i\cond\bx_i) = \N(\bmu(\bx_i),\bm{\sigma}^2_i\bSigma)\label{sample_assumption}
\end{equation}
where, similar to the traditional VAE framework, $\bmu(\bx_i)$ denotes the mean output for the $i$-th sample from the encoder. Unlike the traditional VAE, we assume that the latent variables also have a correlation structure that is proportional to what is mapped by the encoder scaled by a sample-specific variance component parameter.  One could be fully bayesian and also place priors on $\bm{\beta}$ and $\bm{\Sigma}$; however, to keep the simple structure of original VAEs and save computational time, we derive empirical estimators for $\bm{\beta}$ and $\bm{\Sigma}$ simply using a batch of $m$ training samples.

\subsection{Model Training via Variational Inference}

Our goal is to find the ideal trade-off between the reconstruction error and regularization term in the VAE framework loss function. Similar to the logic posed within the WAE, we propose maximizing the lower bound to the marginal likelihood by imposing the aggregated posterior $q_{\phi}(\bz)$ to match a standard normal distribution instead of regularizing each of the independent conditional posteriors $q_{\phi}(\bm{z}_i\cond \bx_i)$. In this section, we will use variational inference to show that this target can be achieved by minimizing the divergence between $q_\phi(\bmu(\bx_i))$ and the standard normal distribution. We begin with the form of the lower bound within the HEBAE framework.

\begin{theorem}
Minimizing $\mathrm{KL}(q_{\phi}(\bmu(\bx_i)) \,\|\, \mathcal{N}(\bm{0}, \bI))$ is equivalent choosing a general isotropic Gaussian as the prior distribution such that $p_{\theta}(\bz_i) = \mathcal{N}(\bmu(\bx_i), \bm{\sigma}_i\bI)$ with the constraint that $\bbeta^\T\bbeta = \bm{0}$. This yields the lower bound to optimize in the HEBAE framework,
\begin{equation}
\begin{aligned}
\mathcal{L}(\btheta,\bphi;\bx_{1:m}) &= \frac{1}{L} \sum_{l=1}^L\sum_{i=1}^{m}\log\,p_{\theta}(\bx_i\cond\bz_i^{(l)})-\lambda_1\sum_{i=1}^{m}\mathrm{KL}(q_{\phi}(\bz_i\cond\bx_i)\,\|\, p_{\theta}(\bz_i))+ \lambda_2\|\bbeta\|^2\\
&= \frac{1}{L}\sum_{l=1}^L\sum_{i=1}^{m}\log\,p_{\theta}(\bx_i\cond\bz_i^{(l)})-\lambda\sum_{i=1}^{m}\mathrm{KL}(q_{\phi}(\bmu(\bx_i)) \,\|\, \mathcal{N}(\bm{0}, \bI)),\label{hebae_lb}
\end{aligned}
\end{equation}
\end{theorem}
Notice that under the KKT conditions, the constraint in the first half of Eq.~\eq{hebae_lb} can be achieved by incorporating an $L_2$-penalty on $\bbeta$ into the objective \cite{karush1939minima, kuhn2014nonlinear}. One can interpret the weight $\lambda$ outside the KL term in the second half of Eq.~\eq{hebae_lb} as a regularization parameter similar to the VAE framework in \eq{elbo2}. The full derivation of the HEBEA lower bound in Eq.~\eq{hebae_lb} can be found in the appendix. Our framework also employs the multivariate reparameterization trick, which yields
\begin{equation}
\bz_i^{(l)} = \bmu(\bx_i)+\sigma_i\odot\bR\bvarepsilon^{(l)}, \quad \quad \bvarepsilon\sim\N(\bm{0},\bI)\label{trick2}
\end{equation}
with $\bm{\Sigma} = \bR\bR^{\T}$ dervied from the Cholesky decomposition of the covariance matrix, where $\bR$ is a lower triangular matrix with real and positive diagonal entries. The variational inference algorithm will impose that $\bR \rightarrow\bI$. In traditional VAEs, there are two key issues. First, $\bR = \bI$ exactly because of the assumed independence among the latent $\bz$. Second, there exists a conflicting issue in the framework where the standard normal priors will impose that $\bsigma^2_i \rightarrow \bm{1}$, while a typical reconstruction loss of the form $\|g(\bz)-\bx\|^2 = \|g(\bmu(\bx)+\bsigma\odot\bvarepsilon)-\bx\|^2$ will force these parameters to tend toward zero. As a result, there is a need to balance the weight between the reconstruction loss and the KL term in the VAE model. However, in the HEBAE framework, since we regularize according to $q_{\phi}(\bmu(\bx_i))$ in Eq.~\eq{gp_prior} which does not depend on any variance component hyperparameters, thus the value of each $\sigma_i^2$ are fully dictated by the reconstruction.

%it's obvious that as the random sampler $\bm{\epsilon}^{(l)}$ is coupled with the $\bm{\sigma}_i$, the reconstruction loss will not be small unless $\bm{\sigma}_i \to 0$.

\subsection{Model Interpretations and Theoretical Comparisons}

In this section, we give intuition behind why the HEBAE framework is better able to avoid posterior collapse and improve upon the performance of the WAE framework. Under the hierarchical model assumptions in Eqs.~\eq{gp_prior} and \eq{sample_assumption}, the aggregated posterior is
\begin{equation}
q_\phi(\bz) = \int q_\phi(\bz\cond\bx_i) p_d(\bx_i) \diff \bx_i \approx \N(\bbeta,[1+\mathcal{U}(\bm{\sigma}^2)]\bSigma)\label{marginal_z}
\end{equation}
where $\mathcal{U}(\bm{\sigma}^2)$ is an averaged estimate of the variance component over all $m$ samples in the training batch. The objective of Eq.~\eq{hebae_lb} will impose that $\bSigma\rightarrow\bI$ and the reconstruction loss will force $\mathcal{U}(\bm{\sigma}^2)$ goes to small. In this optimal case,  $q_\phi(\bz) \approx q_\phi(\bmu(\bx_i))$ and converge to a standard normal distribution. The reason that we pursue this approximate aggregated posterior by incorporating variational inference instead of deterministic updates is that it provides a natural probabilistic way for sampling from the conditional posterior and injecting noise into the decoder so as to smooth out the latent space. More specifically, the variational inference framework enables the HEBAE to pursue a tighter bound than the WAE. As previously shown \cite{hoffman2016elbo}, the ELBO for the VAE can be rewritten as
\begin{equation}
\begin{aligned}
\mathcal{L}_{\text{VAE}}(\btheta,\bphi;\bx_{1:m})= &\frac{1}{L}\sum_{l=1}^L\sum_{i=1}^{m}\log\,p_{\theta}(\bx_i\cond\bz_i^{(l)})-\lambda\sum_{i=1}^{m}\mathrm{KL}(q_{\phi}(\bz)\,\|\, p_{\theta}(\bz))\\
&-\left\{m\,\log\,N-\sum_{i=1}^{m}\mathbb{E}_{q_{\phi}(\bz)}\left[\mathbb{H}[q_\phi(i\cond\bz)]\right]\right\}\label{hebae_lb2}
\end{aligned}
\end{equation}
where the first term is the reconstruction; the second term is the KL divergence between the aggregated posterior and the standard normal prior; and the third term is the ``index-code mutual information'' where $\mathbb{H}[q_{\phi}(i\cond\bz_i)] = \E[\log\,q_{\phi}(i\cond\bz_i)]$ denotes the entropy of variational posterior over the sample indices in the training batch. In the WAE setting, the objective only has the reconstruction and a regularizer that minimizes the distance between the aggregated posterior and the standard normal distribution. Thus, as stated in the original paper, the difference between WAE and VAE can be viewed as dropping the mutual information term from the objective, which is effectively pursuing a ``looser'' lower bound. Because we maintain the variational inference set up in Eq.~\eq{hebae_lb}, our model lies between the traditional VAE and WAE. To see the advantages of HEBAE, we now provide the closed form of our objective Eq.~\eq{hebae_lb} as the following

\begin{equation}
\begin{aligned}
\mathcal{L}_{\text{HEBAE}}(\btheta,\bphi;\bx_{1:m})\approx \frac{1}{L}\sum_{l=1}^L\sum_{i=1}^{m}\log\,p_{\theta}(\bx_i\cond\bz_i^{(l)})-\lambda\left[\tr(\bm{\Sigma})-\textnormal{k} - \log\,|\bSigma|-\bbeta^{\T}\bbeta\right].\\
\label{hebae_lb3}
\end{aligned}
\end{equation}
There are three main benefits from our choice to maximize the above. The first directly improves upon the VAE framework. Notice that the the sample specific variance components $\sigma^2_i$ do not influence the KL divergence terms which can effectively help balance the trade-off between the reconstruction and regularized loss functions. By comparison, in traditional VAEs, these variance components will be pushed towards one which is against the goal of the reconstruction. This creates a conflict as one needs to creatively balance the objectives between the two loss functions. As discussed earlier, this benefit is also related to the goal in WAE that aims to match the aggregated posterior to standard normal prior and drop the mutual information from the objective. Indeed, previous work has shown that by accounting for the mutual information term in VAEs, one can prevent over-regularization and posterior collapse \cite{he2019lagging, Phuong2018TheMA}.

The second benefit is that our hierarchical assumption with an empirical estimator enables direct penalization of the covariance matrix $\bm{\Sigma}$ to converge onto the identity matrix $\bI$ (see Eq.~\eq{elbo2} versus Eq.~\eq{hebae_lb3}). Thus, our posterior is expected to be more consistent with the independence assumption compared to a traditional VAE.

The final benefit is based upon improving the deterministic approach in the WAE framework. It is clear that the mutual information is bounded between $[0, \log\,N]$. In WAE framework, the index $i$ and latent variable $\bz$ have a deterministic relationship such that $q(i\cond\bz)$ is set to be a delta function. This will result in the entropy $\mathbb{E}_{q_{\phi}(\bz)}\left[\mathbb{H}[q_\phi(i\cond\bz)]\right] = 0$ and the mutual information is fixed at $\log\,N$. Though this does help prevent over-regularization in autoencoders and improves performance when generating new data compared to VAEs, completely removing the mutual information from the objective is maximizing a looser bound on the marginal likelihood. Our framework, on the other hand, enables adaptive learning of each $\sigma^2_i$ based on the reconstruction. Since the resulting estimate of $\sigma^2_i$ in the HEBAE framework is not exactly 0, the index-code mutual information will be large enough to prevent over-regularization (i.e., close to $\log\,N$) but will simultaneously introduce enough uncertainty to $q(i\cond\bz)$ such that we are able to sample and estimate a smooth latent space.

\section{Experiments}

In this section, we empirically assess our model against the traditional VAE and WAE frameworks. Our aim is to evaluate the following metrics: \textit{(i)} whether our hierarchical framework can better balance the trade-off between the regularization and reconstruction losses than the traditional VAE; \textit{(ii)} whether our variational posteriors better converge onto the standard normal prior assumptions; and \textit{(iii)} whether our variational inference algorithm results in a decoder that can generate better quality samples than VAEs and WAEs.

 \begin{figure}[hbt!]
\centering
\includegraphics[width=\textwidth]{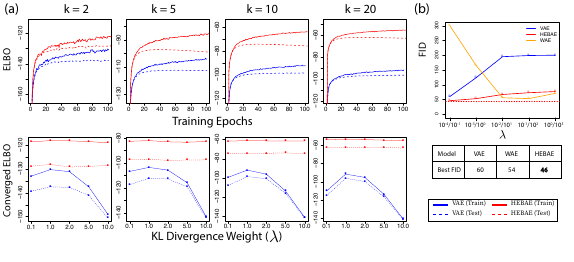}
\caption{HEBAE outperforms VAE and WAE on all three metrics measured. (a) Top row shows that the ELBO of HEBAE converges faster to a better optimum than VAE in all experiments with different latent dimension $k$. Bottom row shows that HEBAE is less sensitive to different KL divergence weights ($\lambda$) while VAEs are susceptible to over-regularization. Results are based on the MNIST dataset. (b) Comparison of FID scores for HEBAE, VAE, and WAE on the CelebA dataset. HEBAE is less sensitive to $\lambda$ and has the lowest FID score.}
\label{fig:elbos}
\end{figure}
\begin{figure}[hbt!]
    \centering
    \includegraphics[width=\textwidth]{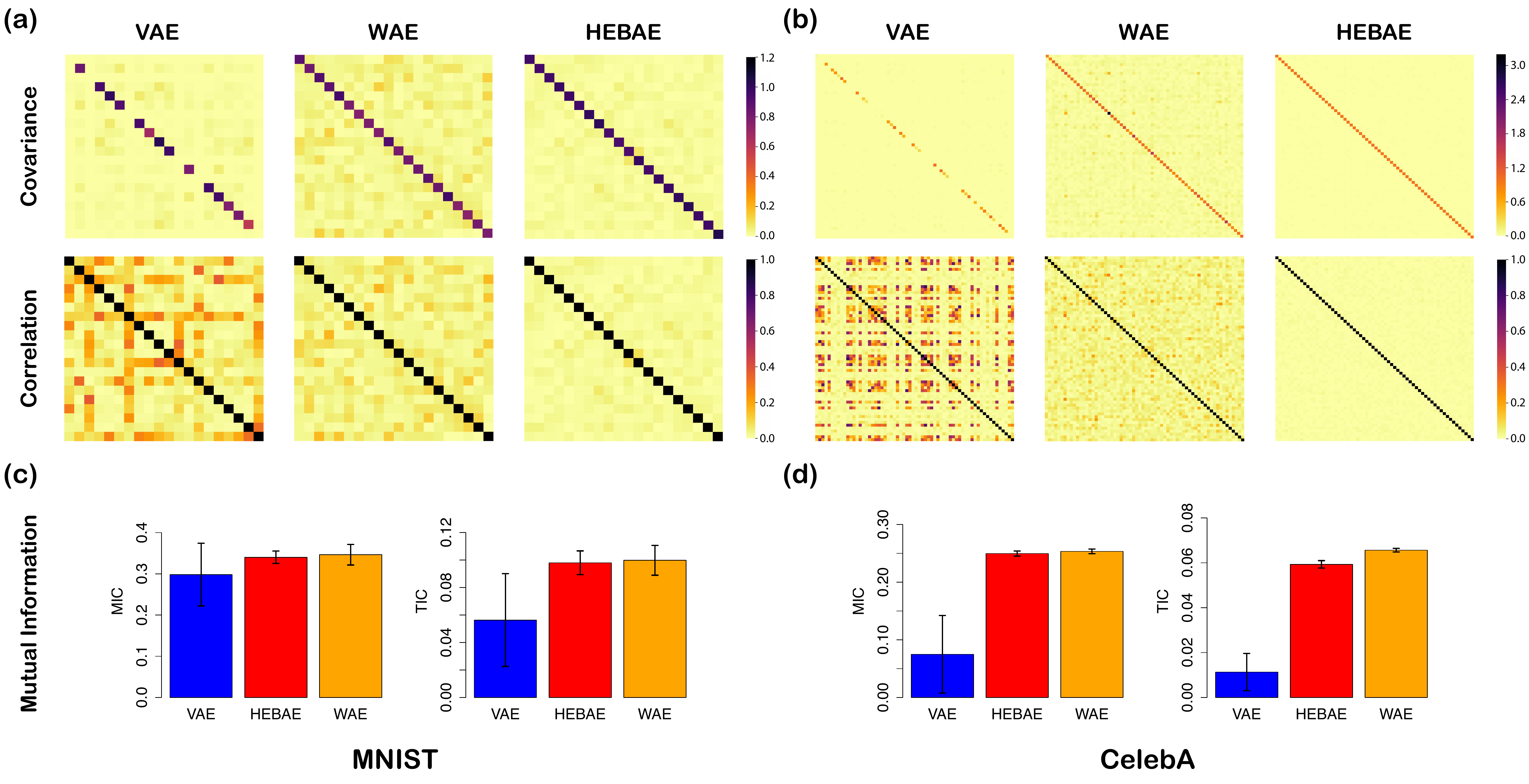}
    \caption{ The estimated posterior of the HEBAE framework is more consistent with the standard normal prior compared to the VAE and WAE frameworks, in both MNIST and CelebA analyses.
     (a, b) Top row shows the absolute value of the variance-covariance matrices. Bottom row shows the correlation matrices. Results are based on MNIST dataset. (c, d) Averaged mutual information measurements: Maximal Information Coefficient (MIC) and Total Information Coefficient (TIC) \cite{albanese2013minerva} computed using index $i$ and each dimension of latent variable $\bz$. HEBAE maintains higher mutual information than VAEs, but slightly smaller than the WAE.}
    \label{fig:matching}
\end{figure}

We evaluate each approach using two datasets: MNIST \cite{726791} and CelebA \cite{liu2015faceattributes}. The MNIST dataset contains 60,000 training images and 10,000 test images, while CelebA contains 202,599 images in total. For MNIST, all models were trained using the same simple two dense layer architectures for both the encoder and decoder. We then carried out experiments with different $k$-dimensions for the latent variables $\bm{z}$. Specifically, we examine the common values $k = \{2, 5, 10, 20\}$ \cite{kingma2013autoencoding, tolstikhin2017wasserstein, ghosh2019variational} and report the results for $k=20$ in the main text. For the CelebA analyses, we adopted the convolution architectures from Tolstikhin et al.~(2017) \cite{tolstikhin2017wasserstein} for all the models with the common choice $k = 64$-dimensional space for the latent variables \cite{tolstikhin2017wasserstein, ghosh2019variational}. All the models are trained with an Adam optimizer \cite{kingma2014adam}. More details about architectures and training procedures could be found in Appendix B.

Since HEBAE utilizes variational inference, we can directly compare its maximized ELBO with the ELBO from a VAE. In Fig.~\ref{fig:elbos}a, we plot these ELBOs against training epochs for different dimensionality of $k$ in the MNIST data and find that our model consistently converges to a higher ELBO faster than VAEs through all experiments. Next, we plot the maximized ELBO against various weight parameter values ($\lambda$) for the KL loss (see the bottom panels of Fig.~\ref{fig:elbos}a). VAEs are generally sensitive to the choice of the KL weight and can easily over-regularize which leads to a much lower ELBO. Our model, on the other hand, is much less sensitive. To further investigate model sensitivity to hyperparameter tuning, in the CelebA dataset, we vary either the KL weight (in the VAE and HEBAE) or the MMD weight (in the WAE) and plot the converged $\textit{Fr\'echet Inception Distance}$ (FID) introduced by ~\cite{heusel2017gans} in Fig.~\ref{fig:elbos}b. As the weight parameter gets closer to 0, each model should behave like a regular autoenocoder and performs worse in all three frameworks (i.e., have a higher FID score). Again, we can see that HEBAE is generally less sensitive to the regularize loss weight. 

Next, we investigate how well the variational posterior distribution of the latent variables converges onto the standard normal within the context of each of these three frameworks. In Fig.~\ref{fig:matching}, we plot the variance-covariance and correlation matrices at convergence for both the MNIST ($k=20$) and CelebA ($k=64$) datasets. Namely, at convergence, a successful model will have an independent covariance structure between the latent variables. We can see specific instances where the VAE experiences posterior collapse: this is illustrated on the top row of Fig.~\ref{fig:matching} where the variance terms on the diagonals reduce close to 0. This is a phenomenon that is not experienced in within the HEBAE and WAE frameworks. Altogether, Fig.~\ref{fig:matching} illustrates the point that matching the aggregate posterior can remedy the over-regularization problem. Furthermore, the hierarchical structure in the HEBAE framework allows it to converge onto the target standard normal prior more consistently than VAE and WAE (i.e., see the second and third rows of Fig.~\ref{fig:matching}). Specifically, from the correlation plot, we can directly see that our aggregated posteriors converge onto the independent assumption better than the other two approaches.

\begin{figure}[hbt!]
    \centering
    \includegraphics[width=\textwidth]{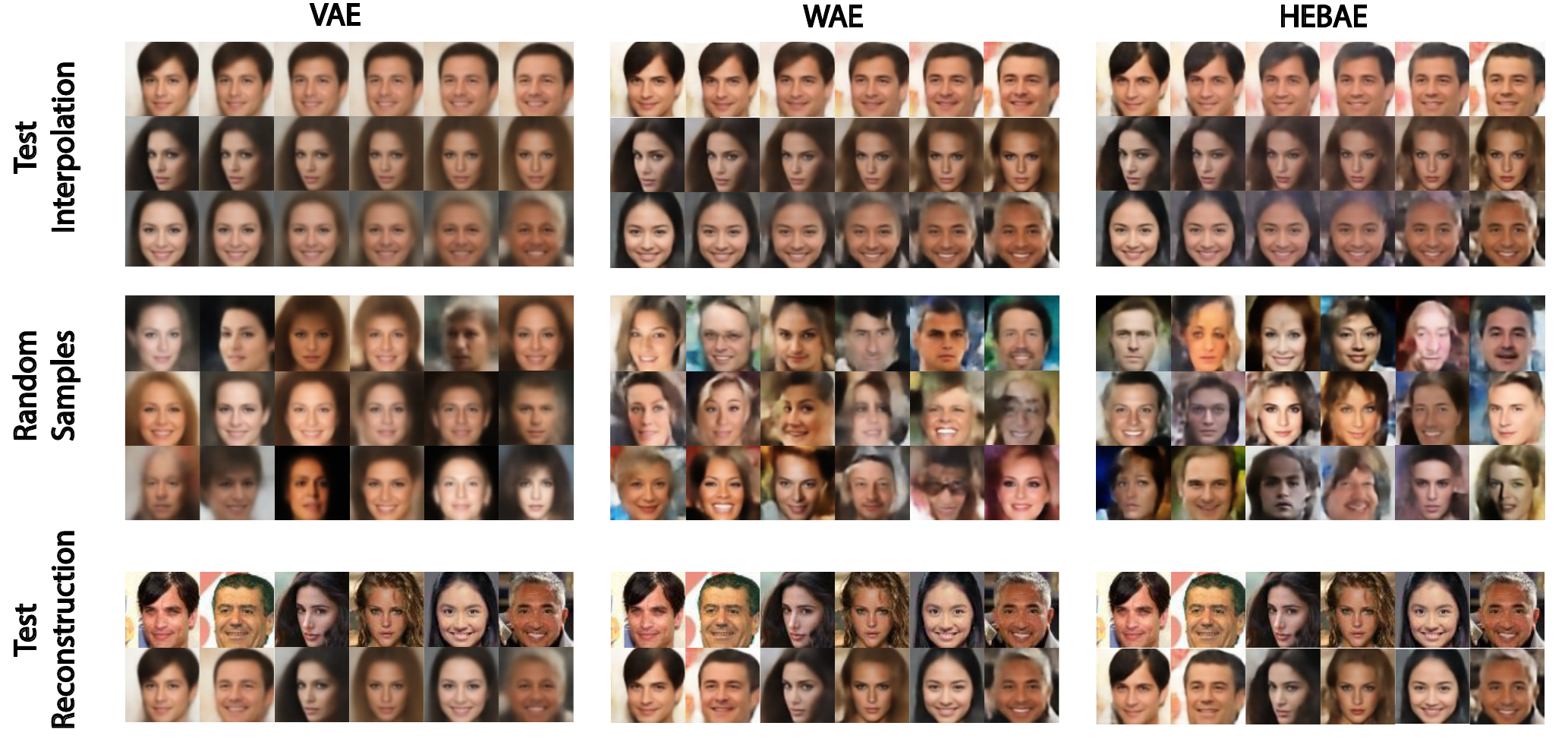}
    \caption{HEBAE produces qualitatively higher-quality images based on the CelebA dataset than the VAE and WAE frameworks. Results on MNIST can be found in the Appendix.}
    \label{fig:generated_images_CelebA}
\end{figure}

Lastly, we demonstrate how HEBAE benefits from turning the optimization goals of a WAE into a probabilistic model like a VAE. 
%The posterior estimates of the variance components $\sigma^2$ in the HEBAE framework are smaller than the variances in VAEs which results in less distributional overlap in the latent space and allows for better reconstruction \cite{tolstikhin2017wasserstein, hoffman2016elbo}. Still having small-yet-nonzero estimates for each $\sigma^2$ means our model has the ability to smooth the latent space as compared to the deterministic approach in WAEs which automatically sets $\sigma^2=0$. 
Remember that WAE attempt to maximize a ``looser'' lower bound (see discussion around Eq.~\eq{hebae_lb2}). In Fig.~\ref{fig:matching}c and Fig.~\ref{fig:matching}d, we assess the corresponding mutual information between the sample-index $i$ and latent variables $\bz$ at the convergence. As WAEs do not provide a closed-form ELBO, we follow previous work \cite{albanese2013minerva} and approximate mutual information using the average Maximal Information Coefficient (MIC) and Total Information Coefficient (TIC) for different dimensions of $\bz$. Here, we see that the VAE has much smaller mutual information, while the WAE has the highest scores. The mutual information from the HEBAE sits between the VAE and WAE, respectively. Similar to previous work, Maintaining a high mutual information prevents HEBAE from over-regularization and posterior collapse  \cite{tolstikhin2017wasserstein, Phuong2018TheMA}. Also since we do this in a probabilistic framework, our model is able to sample posterior estimates and smooth out the latent space which improves sample qualities compared with WAE \cite{ hoffman2016elbo}. To see this, we assess the quality of samples generated by HEBAE. In Fig.~\ref{fig:elbos}b and Fig.~\ref{fig:generated_images_CelebA}, we provide qualitative comparisons using test reconstruction, test interpolation, and random generated samples for the CelebA data. Quantitatively, HEBAE shows to have lower FID scores (with the lowest score being 46). Results for MNIST can be found in Appendix C.

{}

% in  The generated samples Fig.\ref{fig:generated_images_CelebA} and Fig.\ref{fig:generated_images_MNIST} and [FID tables] show that this small-yet-nonzero variance give us better generated images. (The contributions could also be part of the independent z as well. We have experiments where we don't sample from conditional posteriors but the performance is worse so the smoother latent space is definitely helpful here. I could add that as well if it's necessary.) More importantly, from Fig.\ref{fig:MI_MNIST} and [MI figure for CelebA], we empirically show that the smaller the variances for conditional posterior, the larger the mutual information. This correspond to the theory in the method section and many other [citations] as well that is, we keep MI to be high (smaller variances for conditional posteriors) but not the maximum (compltely dropped from the loss like in WAEs so it's not minimized at all and remain at maximum), we are technically pursuing an tighter ELBO (with constraint though) which gives better sample qualities. Sample qualities are evaluated quantitatively by all FID results and qualitatively by Fig.\ref{fig:generated_images_CelebA} and Fig.\ref{fig:generated_images_MNIST}.

\section{Conclusion}

In this work, we present Hierarchical Empirical Bayes Autoencoder (HEBAE), a new framework for probabilistic generative modeling. Our theoretical work connects the probabilistic framework of the VAE and the deterministic objective of the WAE. We illustrated the trade-off between reconstruction of samples and regularization of the latent space in the VAE and have shown how HEBAE can prevent over-regularization. We further demonstrate that matching the posterior to a more general prior distribution avoids issues with posterior collapse. 
%We also compare our variational inference framework to the deterministic estimation done in WAE.
In experiments assessing mutual information, it is clear that the sampling mechanism and probabilistic priors of HEBAE yield a smoother latent space to connect the encoder and decoder.
The hierarchical assumption present in HEBAE leads to less sensitivity to initial settings of hyperparameters and enables fast algorithmic convergence.
%While the hiearchy of our model provides tangible benefits, we do not take a fully Bayesian approach to learn posterior parameters, and expect a fully Bayesian approach might lead to even stronger empirical results.
It remains to disentangle the precise contribution of performance in our model between estimating an aggregated posterior and the form of the prior itself -- which yields a latent space with orthogonal components that more strictly conform to mean-field assumptions.
%On this note, while theoretical results have connected particular forms of linear auto-encoders to principal components analysis (PCA), it remains to understand how our non-linear auto-encoder with an independence assumption on the latent space could be related to non-linear variants of PCA.
%Furthermore, we have shown how the hierarchical assumption of HEBAE better enables algorithmic convergence.
Finally, we validate our assumptions and theories with empirical experiments and show that HEBAE can generate higher quality images than its auto-encoder counterparts. GANs still remain state-of-the-art in generative modeling for images, and we expect insights from our probabilistic framework could be transferred into further improvements in GAN architecture.

\section*{Broader Impact}

In this work, we propose a new method named HEBAE for generative modeling. We show that by combining the merits of probabilistic and deterministic autoencoders, we can develop an easily trainable method that generates higher quality data than the state-of-the-art. It is true that we could focus on the negative broader impacts of our work which might include generating more realistic fraudulent images and fake news stories. However, while we acknowledge these possibilities, we will instead focus on the positive. Naturally, HEBAE will benefit the many communities that are seeking methods for photograph generation, attribute interpolation, image-to-image translation, etc. However, our work also has the potential to be used in broader machine learning applications such as semi-supervised and disentanglement learning, as well as within other scientific fields such as natural language processing (NLP), robotics, and genomics. Indeed, one merit of our method is its ability to prevent over-regularization and posterior collapse in traditional variational autoencoders. For example, in NLP, the widely used autoregressive decoder is known to suffer from these same issues. The genomics community is one place that stand to gain great benefit from our new approach. Generating realistic artificial genomes can be used to improve human genome privacy issues and, in rare disease studies where data are far less prevalent, generative models can have a massive impact in helping us derive successful treatment strategies for future patients we have yet to observe. Furthermore, HEBAE is built based on an autoencoder structure which we showed can generate independent lower dimensional representations of data similar to non-linear principal component analysis (PCA); thus, one can also use our method for efficient dimension reduction purposed in studies that aim to analyze high-throughput sequencing assays.

\section*{Acknowledgements}

This research was conducted using computational resources and services at the Center for Computation and Visualization (CCV), Brown University. This work was supported by grants from the US National Institutes of Health (R01 GM118652, P20GM109035, P20GM103645, and 2U10CA180794-06), the National Science Foundation (CAREER award DBI-1452622), and an Alfred P. Sloan Research Fellowship. Any opinions, findings, and conclusions or recommendations expressed in this material are those of the authors and do not necessarily reflect the views of any of the funders or supporters.

\clearpage
\newpage
\appendix

\setcounter{theorem}{0}

\section*{Appendix}

\section{Derivation of Theorem 1}

In the main text, we show that the main advantage of the Hierarchical Empirical Bayes Autoencoder (HEBAE) framework is having the ability to find the ideal trade-off between the reconstruction error and regularization term in the traditional variational autoencoder (VAE) loss function. To do so, we propose maximizing the evidence lower bound (ELBO) with respect to the aggregated posterior distribution $q_\phi(\bz)$. In other words, instead of regularizing each independent conditional posterior $q_\phi(\bz_i\cond\bx_i)$ during training, HEBAE instead imposes that $q_\phi(\bz)$ matches a standard normal distribution. This led to the following statement about the closed-form for the lower bound within the HEBAE framework.
\begin{theorem}
Minimizing the Kullback-Leibler divergence $\mathrm{KL}(q_{\phi}(\bmu(\bx_i)) \,\|\, \mathcal{N}(\bm{0}, \bI))$ is equivalent choosing a general isotropic Gaussian as the prior distribution such that $p_{\theta}(\bz_i) = \mathcal{N}(\bmu(\bx_i), \sigma^2_i\bI)$ with the constraint that $\bbeta^\T\bbeta = \bm{0}$. This yields the lower bound to optimize in the HEBAE framework,
\begin{align}
\mathcal{L}(\btheta,\bphi;\bx_{1:m}) &= \frac{1}{L} \sum_{l=1}^L\sum_{i=1}^{m}\log\,p_{\theta}(\bx_i\cond\bz_i^{(l)})-\lambda_1\sum_{i=1}^{m}\mathrm{KL}(q_{\phi}(\bz_i\cond\bx_i)\,\|\, p_{\theta}(\bz_i))+ \lambda_2\|\bbeta\|^2\label{step1}\\
&= \frac{1}{L}\sum_{l=1}^L\sum_{i=1}^{m}\log\,p_{\theta}(\bx_i\cond\bz_i^{(l)})-\lambda\sum_{i=1}^{m}\mathrm{KL}(q_{\phi}(\bmu(\bx_i)) \,\|\, \mathcal{N}(\bm{0}, \bI)),\label{step2}
\end{align}
\end{theorem}
where we use the following multivariate reparameterization trick
\begin{equation}
\bz_i^{(l)} = \bmu(\bx_i)+\sigma_i\odot\bR\bvarepsilon^{(l)}, \quad \quad \bvarepsilon\sim\N(\bm{0},\bI)\label{trick2mvn}
\end{equation}
with $\bm{\Sigma} = \bR\bR^{\T}$ dervied from the Cholesky decomposition of the covariance matrix between the latent $\bz$ variables and $\bR$ is a lower triangular matrix with real and positive diagonal entries. Notice that both Eq.~\eq{step1} and Eq.~\eq{step2} have the same reconstruction loss as the first term. Therefore, in this section, we will focus on the KL divergence terms and their relationship with the extra constraint. Under KKT conditions, the constraint in Eq.~\eq{step1} can be achieved by incorporating an $L_2$-penalty on $\bbeta$ into the objective \cite{karush1939minima, kuhn2014nonlinear}. One can interpret the weight $\lambda_1$ outside the KL term in Eq.~\eq{step1} as a regularization parameter similar to the VAE framework (where $\lambda_1 = 1$ in the traditional model). Since both Eq.~\ref{step1} and Eq.~\ref{step2} has the reconstruction loss as the first term, then showing their equivalence simply amounts to deriving the relationship between the KL divergence terms and the extra constraint.
% As one could see that both Eq.~\ref{step1} and Eq.~\ref{step2} has the reconstruction loss as the first term. Thus, we will focus on the KL divergence and the extra constraint term.
% Then,if we choose the prior distributions to be $p_{\theta}(\bz_i) = \mathcal{N}(\bmu(\bx_i), \bm{\sigma}_i\bI)$. We can derive the following objective,
% \begin{equation}
% \begin{aligned}
% \mathcal{L}(\btheta,\bphi;\bx_{1:m}) &= \frac{1}{L} \sum_{l=1}^L\sum_{i=1}^{m}\log\,p_{\theta}(\bx_i\cond\bz_i^{(l)})-\lambda_1\sum_{i=1}^{m}\mathrm{KL}(q_{\phi}(\bz_i\cond\bx_i)\,\|\, \mathcal{N}(\bmu(\bx_i), \bm{\sigma}_i\bI))
% \end{aligned}
% \end{equation}
% Notice that we introduce a weight parameter $\lambda_1$ for the KL Divergence loss into the objective as usual while the original ELBO will have $\lambda_1 = 1$. Then, under the KKT condition, the extra constraint $\bbeta^\T\bbeta = \bm{0}$ can be achieved by including a penalty on $\bbeta$, which gives the objective in Eq. ~\ref{step1}.
To begin, we first restate the hierarchical variational family assumption within the HEBAE framework,
\begin{equation}
\begin{aligned}
q_{\phi}(\bz_i\cond\bx_i)\sim \mathcal{N}(\bmu(\bx_i), \sigma^2_i\bSigma), \quad \quad q_{\phi}(\bmu(\bx_i))\sim \mathcal{N}( \bbeta, \bSigma).
\end{aligned}
\end{equation}
Under this model, we can find a closed form expression for the KL term in the objective in Eq.~\eq{step1}
\begin{equation*}
\begin{aligned}
\mathrm{KL}(q_{\phi}(\bz_i\cond\bx_i)\,\|\, p_{\theta}(\bz_i))= \mathrm{KL}(\mathcal{N}(\bmu(\bx_i), \sigma^2_i\bSigma)\,\|\, \N(\bm{0},\sigma^2_i\bI))
\end{aligned}
\end{equation*}
Taking the KL divergence between two Gaussian distributions with some algebraic rearrangement and simplification yields 
\begin{equation}
\begin{aligned}
\mathrm{KL}(\mathcal{N}(\bmu(\bx_i), \sigma^2_i\bSigma)\,\|\, \N(\bm{0},\sigma^2_i\bI))=\tr(\bSigma) - k - \log|\bSigma|\label{Eq5}
\end{aligned}    
\end{equation}
We can plug Eq.~\eq{Eq5} into the objective in Eq.~\eq{step1} to find
\begin{equation}
\begin{aligned}
\mathcal{L}(\btheta,\bphi;\bx_{1:m}) &= \frac{1}{L} \sum_{l=1}^L\sum_{i=1}^{m}\log\,p_{\theta}(\bx_i\cond\bz_i^{(l)})-\lambda_1\sum_{i=1}^{m}\mathrm{KL}(q_{\phi}(\bz_i\cond\bx_i)\,\|\, p_{\theta}(\bz_i))+ \lambda_2\|\bbeta\|^2\\
&= \frac{1}{L} \sum_{l=1}^L\sum_{i=1}^{m}\log\,p_{\theta}(\bx_i\cond\bz_i^{(l)})-\lambda_1\sum_{i=1}^{m}\bigg[\tr(\bSigma) - k - \log|\bSigma|\bigg]+ \lambda_2\|\bbeta\|^2\\
&= \frac{1}{L} \sum_{l=1}^L\sum_{i=1}^{m}\log\,p_{\theta}(\bx_i\cond\bz_i^{(l)})-\lambda\sum_{i=1}^{m}\bigg[\tr(\bSigma) - k - \log|\bSigma|+\bbeta^{\T}\bbeta\bigg]\\
&= \frac{1}{L} \sum_{l=1}^L\sum_{i=1}^{m}\log\,p_{\theta}(\bx_i\cond\bz_i^{(l)})-\lambda\sum_{i=1}^{m}\mathrm{KL}(q_{\phi}(\bmu(\bx_i)) \,\|\, \mathcal{N}(\bm{0}, \bI))
\end{aligned}
\end{equation}
In the settings where $\lambda_1 = \lambda_2$, then the objectives in Eq.~\eq{step1} is equal to the objective in Eq.~\eq{step2} which concludes the proof.

\section{Details on Experiments Setup}

\subsection{MNIST Dataset}

We used the following simple dense architectures for the HEBAE, VAE and WAE models. Note that $k$ denotes the dimension for latent variable $\bm{z}$ and $\textnormal{FC}_{k \times p}$ represents the fully connected layer. Lastly, $p = 2$ for VAE and HEBAE as both need to output variance component terms, while $p = 1$ for WAE. The encoder architectures are then:
\begin{center}
\begin{tabular}{l l} 
\hline
 $\bx \in \mathcal{R}^{784}$ & $\to$ $\textnormal{FC}_{784}$ $\to$ $\textnormal{ReLU}$ \\
[1.5pt]\hline
 \ & $\to$ $\textnormal{FC}_{800}$ $\to$ $\textnormal{ReLU}$ $\to$ $\textnormal{FC}_{k \times p}$ \\
[1.5pt]\hline
%   \ & $\to$ $\textnormal{FC}_{k}$ \quad \\
%  \hline
\end{tabular}
\end{center}
For the architecture in the decoder, we use:
\begin{center}
\begin{tabular}{l l} 
\hline
 $\bz \in \mathcal{R}^{k}$ & $\to$ $\textnormal{FC}_{800}$ $\to$ $\textnormal{ReLU}$ \\
[1.5pt]\hline
  \ & $\to$ $\textnormal{FC}_{800}$ $\to$ $\textnormal{ReLU}$ $\to$ $\textnormal{FC}_{784}$ \\
[1.5pt]\hline
%   \ & $\to$ $\textnormal{FC}_{k}$ \quad \\
%  \hline
\end{tabular}
\end{center}
We used mini-batches with size = 128 and all the models were trained for 100 epochs. The default KL weight $\lambda$ is set to be 1 for VAEs except for the experiments used to generate the bottom row of Fig.~1(a) in the main text where we evaluated each method based on a grid of $\lambda$ values. For the WAE, we adopted a suggestion from Tolstikhin et al.~(2017) \cite{tolstikhin2017wasserstein} and used $\lambda = 10$ for the MMD penalty. We used the Adam optimizer \cite{kingma2014adam} with an initial learning rate of 0.001 and then the learning rate decays at a rate of 0.995 with every epoch. 

\subsection{CelebA Dataset}

For the CelebA analyses, we adopted the convolution architectures from Tolstikhin et al.~(2017) \cite{tolstikhin2017wasserstein, ghosh2019variational}. Similarly, iamges are also center cropped and resized to $64\times 64$ resolution. Here, note that $\textnormal{Conv}_n$ represents the convolution layer with $n$ filters and $\textnormal{ConvT}_n$ represents the transpose convolution layer with $n$ filters. All convolution and transpose convolution layers have filter sizes of $5 \times 5$ with a stride of size 2, except for the last transpose convolution layer of the decoder which has a stride of size 1. Once again, $\textnormal{FC}_{k \times p}$ denotes the fully connected layer where $p = 2$ for VAE and HEBAE and and $p = 1$ for WAE. The encoder architectures are:
\begin{center}
\begin{tabular}{l l} 
\hline
 $\bx \in \mathcal{R}^{64 \times 64 \times 3}$ & $\to$ $\textnormal{Conv}_{128}$ $\to \textnormal{BN}   \to \textnormal{ReLU}$ \\
[1.5pt]\hline
  \ & $\to$ $\textnormal{Conv}_{256}$ $\to \textnormal{BN}   \to \textnormal{ReLU}$  \\
[1.5pt]\hline
  \ & $\to$ $\textnormal{Conv}_{512}$ $\to \textnormal{BN}   \to \textnormal{ReLU}$  \\
[1.5pt]\hline
   \ & $\to$ $\textnormal{Conv}_{1024}$ $\to \textnormal{BN}   \to \textnormal{ReLU} $  \\
[1.5pt]\hline
    \ &  $\to \textnormal{FLATTEN}   \to \textnormal{FC}_{64\times p}$  \\
[1.5pt]\hline
%   \ & $\to$ $\textnormal{FC}_{k}$ \quad \\
%  \hline
\end{tabular}
\end{center}
The decoder architectures are:
\begin{center}
\begin{tabular}{l l} 
\hline
 $\bz \in \mathcal{R}^{64}$ & $\to$ $\textnormal{FC}_{8 \times 8 \times 1024}$ \\
[1.5pt]\hline
  \ & $\to$ $\textnormal{ConvT}_{512}$ $\to \textnormal{BN}   \to \textnormal{ReLU}$  \\
[1.5pt]\hline
  \ & $\to$ $\textnormal{ConvT}_{256}$ $\to \textnormal{BN}   \to \textnormal{ReLU}$  \\
[1.5pt]\hline
   \ & $\to$ $\textnormal{ConvT}_{128}$ $\to \textnormal{BN}   \to \textnormal{ReLU} \to \textnormal{ConvT}_{1}$  \\
[1.5pt]\hline
%   \ & $\to$ $\textnormal{FC}_{k}$ \quad \\
%  \hline
\end{tabular}
\end{center}
Similar to the MNIST analyses, we used mini-batches of size = 100 and all the models are trained up to 100 epochs. We used a learning strategy from Tolstikhin et al.~(2017) \cite{tolstikhin2017wasserstein} where we set that the initial learning rate to be $10^{-4}$ and then was decreased it by a factor of 2 after 30 epochs, by a factor of 5 after 50 epochs, and by a 10 after 70 epochs. The choices of $\lambda$ are shown in main text Fig.~1(b).

\clearpage
\newpage
\section{Sample Results of MNIST}
\begin{figure}[H]
\centering
\includegraphics[width=\textwidth]{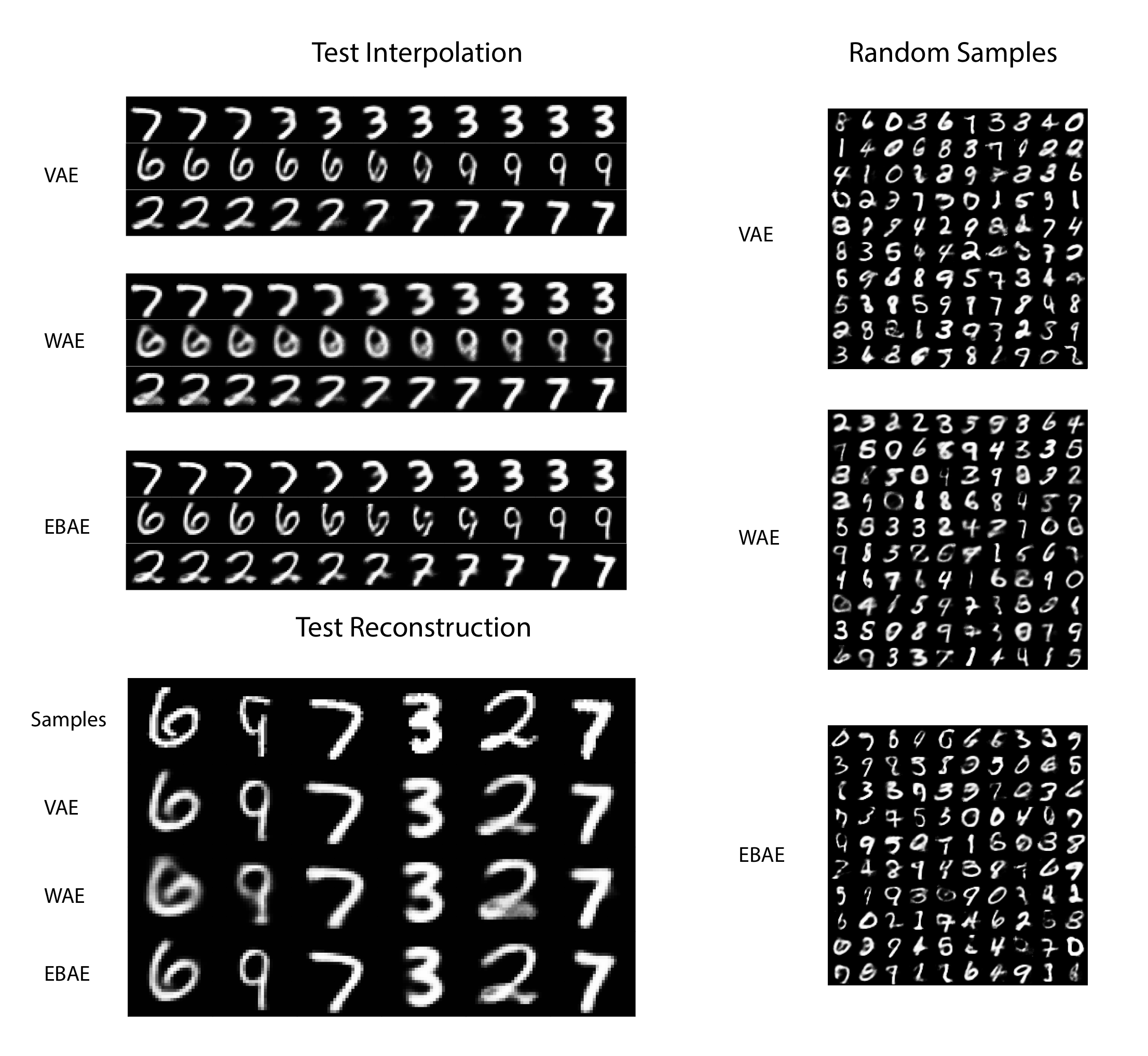}
\caption{HEBAE produces qualitatively higher-quality images than the VAE and WAE frameworks. Samples are generated with $k = 10$, $\lambda = 1$ for VAE and HEBAE, and $\lambda = 10$ for WAE.}
\label{fig:generated_images_MNIST}
\end{figure}
\bibliography{neurips_2020.bib}
\bibliographystyle{plos}

\end{document}